\documentclass[preprint,12pt]{elsarticle}

\usepackage{amssymb}
\usepackage{booktabs}

\usepackage{xcolor}
\usepackage{tcolorbox}
\usepackage{amsmath}
\usepackage{geometry}

\journal{journal}
\usepackage{amsmath}
\usepackage{mathrsfs}
\usepackage{hyperref}

\begin{document}

\begin{frontmatter}



\title{Granular Ball Twin Support Vector Machine\\
with Universum Data}


\author[inst2]{M. A. Ganaie \corref{cor1}}
\ead{mudasir@iitrpr.ac.in}
\cortext[cor1]{Corresponding author}

\author[inst2]{Vrushank Ahire}
\ead{
2022csb1002@iitrpr.ac.in}
\affiliation[inst2]{organization={Department of Computer Science and Engineering, Indian Institute of Technology Ropar},
            city={Rupnagar},
            postcode={140001}, 
            state={Punjab},
            country={India}}
\begin{abstract}

Classification with support vector machines (SVM) often suffers from limited performance when relying solely on labeled data from target classes and is sensitive to noise and outliers. Incorporating prior knowledge from Universum data and more robust data representations can enhance accuracy and efficiency. Motivated by these findings, we propose a novel Granular Ball Twin Support Vector Machine with Universum Data (GBU-TSVM) that extends the TSVM framework to leverage both Universum samples and granular ball computing during model training. Unlike existing TSVM methods, the proposed GBU-TSVM represents data instances as hyper-balls rather than points in the feature space. This innovative approach improves the model’s robustness and efficiency, particularly in handling noisy and large datasets. By grouping data points into granular balls, the model achieves superior computational efficiency, increased noise resistance, and enhanced interpretability. Additionally, the inclusion of Universum data, which consists of samples that are not strictly from the target classes, further refines the classification boundaries. This integration enriches the model with contextual information, refining classification boundaries and boosting overall accuracy. Experimental results on UCI benchmark datasets demonstrate that the GBU-TSVM outperforms existing TSVM models in both accuracy and computational efficiency. These findings highlight the potential of the GBU-TSVM model in setting a new standard in data representation and classification.
\end{abstract}



\begin{center}
    \textbf{\Large Granular Ball Twin Support Vector Machine \\
with Universum Data}
\end{center}

\vspace{0.8cm}

\begin{tcolorbox}[colback=blue!5!white, colframe=blue!75!black, title=\textbf{Research Highlights}]
\vspace{0.3cm}
\begin{itemize}
    \item \textbf{\textcolor{blue}{Innovative Data Representation with Granular Balls:}} \\
    The GBU-TSVM model employs an innovative approach by representing data instances as granular balls rather than conventional points. This method improves the model's robustness and efficiency, especially in handling noisy and large datasets. By grouping data points into granular balls, the model achieves better computational efficiency, increased noise resistance, and enhanced interpretability, establishing a new standard in data representation.
    \vspace{0.3cm}
    \item \textbf{\textcolor{blue}{Enhanced Generalization using Universum Data :}} \\
    The GBU-TSVM incorporates Universum data, which includes samples outside the target classes, to significantly improve generalization capabilities. This integration enriches the model with contextual information, refining classification boundaries and increasing overall accuracy. Universum data enables the classifier to perform better on benchmark datasets, demonstrating the model's ability to utilize additional knowledge for more precise predictions.
    \vspace{0.3cm}
    \item \textbf{\textcolor{blue}{Refined Learning with Modified Hinge Loss Function:}} \\
    The model includes an advanced hinge loss function that accounts for the radii of granular balls, leading to a more accurate error measure and learning process. This modification allows for a detailed error assessment, enhancing the model's learning efficiency and decision boundary precision. By addressing the limitations of existing TSVM models, this innovation sets a new benchmark in the field of machine learning classifiers.
\end{itemize}
\vspace{0.3cm}
\end{tcolorbox}

\newpage



\end{frontmatter}



\section{Introduction}
\label{sec:Introduction}

\begin{figure}[h]
  \centering
  \includegraphics[width=0.40\linewidth]{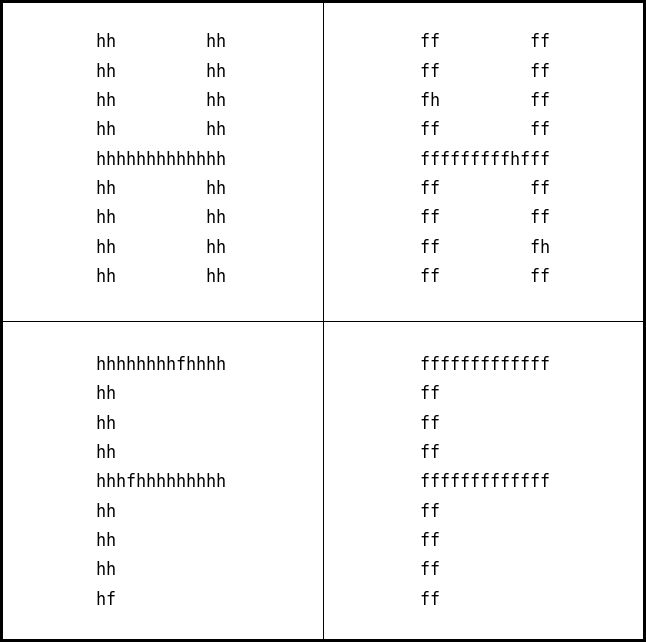}
  \caption{Global precedence in human cognition}
  \label{fig:figure1}
\end{figure}

Over the past few years, classification techniques in machine learning have witnessed remarkable progress, particularly with the advent of Support Vector Machines (SVMs) and their numerous adaptations. Traditional SVMs have gained widespread usage for classification tasks due to their robustness and effectiveness in handling high-dimensional data. However, despite their success, SVMs have certain inherent limitations, such as sensitivity to noise and inefficiency when dealing with large, complex datasets.
Recognizing these limitations, researchers have actively explored novel methods to enhance the performance and capabilities of SVM classifiers. One such approach is Granular Computing, which enables the representation and processing of information in a hierarchical manner, from coarse to fine granules (Xia et al., 2023) \cite{xia2023granular}. This cognitive approach aligns with the way humans intuitively manage complexity (Figure \ref{fig:figure1}), providing a robust framework for data analysis and knowledge discovery. Within this realm, Granular-ball Computing has emerged as a powerful paradigm, involving the partitioning of datasets into hyper-balls of varying sizes, known as granular-balls, which encapsulate groups of data points. These granular-balls serve as an intermediate representation between the raw data points and the final classification model, offering several advantages, including improved computational efficiency, enhanced robustness to noise, and better interpretability.\\
Building upon the principles of Granular-ball Computing, researchers developed the Granular-ball Support Vector Machine (GB-SVM) (Xia et al., 2019)\cite{xia2019granular}, an innovative classifier that utilizes granular-balls as input instead of individual data points, leveraging the inherent benefits of this approach. While GB-SVM addressed some of the limitations of traditional SVMs, it still faced challenges in handling complex datasets with noise and outliers.

Concurrently, the concept of Twin Support Vector Machines (TSVMs) (Jayadeva et al.,(2007) \cite{Jayadeva2007twin}; Kumar et al.,(2008) \cite{kumar2008application}; Shao et al.,(2012) \cite{shao2012coordinate}) gained traction as an efficient alternative to traditional SVMs for binary classification tasks. Unlike SVMs that find a single hyperplane to separate classes, TSVMs construct two non-parallel hyperplanes, each closer to one of the two classes while maintaining a margin from the other. This method simplifies the problem by breaking it down into two smaller and more manageable quadratic programming problems (QPP) tasks, leading to faster training and good generalization performance.

\begin{figure}[h]
  \centering
  \includegraphics[width=0.45\linewidth]{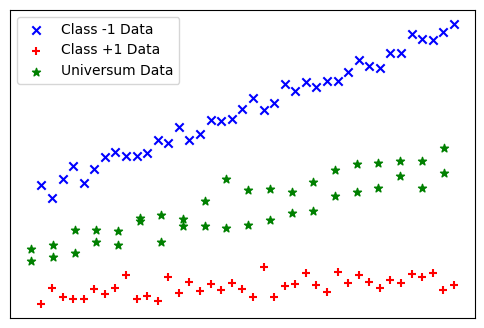}
  \caption{Visualizing Universum Data and Classes}
  \label{fig:figure2}
\end{figure}

To further enhance the capabilities of TSVMs, researchers explored the integration of Universum data into the SVM framework (Weston et al., (2006) \cite{weston2006inference}). Universum data comprises examples that fall outside of the target classes but offer valuable contextual information to enhance the learning process (Figure \ref{fig:figure2}). By incorporating these Universum samples, classifiers can achieve better generalization and accuracy by leveraging prior knowledge about the problem domain.\\
The concept of Universum learning was successfully applied to TSVMs, leading to the development of the Universum TSVM (U-TSVM) (Qi et al., (2012) \cite{qi2012twin}). Unlike traditional U-SVM, U-TSVM utilizes two hinge loss functions to place Universum data in a non-parallel insensitive loss tube, allowing for more flexible exploitation of prior knowledge. This approach demonstrated superior performance compared to traditional SVMs in various tasks.

Recognizing the potential synergies between Universum learning, TSVMs, and Granular-ball Computing, researchers proposed various extensions and improvements to address the limitations of existing methods. Xu et al. (2016a \cite{xu2016least}, 2016b \cite{xu2016nu}) extended the concept of U-TSVM by introducing regularization terms and least-squares formulations, respectively, to improve generalization performance and computational efficiency.

Concurrently, researchers tackled related challenges, such as class imbalance and multi-task learning scenarios. Richhariya et al. (2020a) \cite{richhariya2020reduced} introduced the reduced universum twin support vector machine for class imbalance learning (RUTSVM-CIL), incorporating Universum learning with SVM to address class imbalance problems. They used a small-sized rectangular kernel matrix to reduce computational time, making their method suitable for large-scale imbalanced datasets. Moosaei et al. (2022 \cite{moosaei2022universum}, 2023 \cite{moosaei2023inverse}) proposed the universum parametric-margin $\nu$-support vector machine ($\mathcal{U}$ Par-$\nu$-SVM) and the inverse free reduced universum twin support vector machine for imbalanced data classification, respectively, aiming to improve generalization performance by integrating Universum data and exploiting parametric margins or inverse-free formulations.

Xiao et al. (2021a \cite{xiao2021multi}, 2021b \cite{xiao2021transductive}) extended the application of Universum learning to multi-task and transductive settings, introducing novel methods that leveraged Universum data in these contexts. Kumar et al. (2021) \cite{kumar2021universum} proposed the universum based Lagrangian twin bounded support vector machine (ULTBSVM) for classifying EEG signals, using Universum data to include prior data distribution knowledge.

Building upon these advancements, Liu et al. (2022) \cite{liu2022adaptive} and Li et al. (2022) \cite{li2022improved} developed adaptive robust Adaboost-based twin support vector machines with Universum learning (ARABUTWSVM) and an improved parametric-margin universum TSVM, respectively, aiming to enhance the robustness and generalization performance of existing methods.

Motivated by these challenges and advancements, we propose a novel Granular Ball Twin Support Vector Machine with Universum Data (GBU-TSVM) that synergistically combines the strengths of TSVM, Universum learning, and granular ball computing (Xia, 2022) \cite{xia2022gbsvm} . Our GBU-TSVM extends the TSVM framework by representing data instances as granular balls instead of points in the feature space. This innovative approach improves the model's robustness and efficiency, particularly in handling noisy and large datasets. By grouping data points into granular balls, the model achieves superior computational efficiency, increased noise resistance, and enhanced interpretability. Additionally, the inclusion of Universum data further refines the classification boundaries by enriching the model with contextual information from samples that are not strictly from the target classes (Xiao et al., (2021a)) 
 \cite{xiao2021multi}.
 
The integration of Universum learning and granular ball computing in the GBU-TSVM model addresses several limitations of existing methods.
\renewcommand{\labelitemi}{$\diamond$}
\begin{itemize}
  \item The granular ball representation provides a robust and efficient alternative to traditional point-based representations, mitigating the impact of noise and outliers while reducing computational complexity.
  \item The incorporation of Universum data enhances the model's ability to capture prior knowledge and meaningful concepts from the problem domain, leading to improved classification accuracy.
  \item The twin support vector machine formulation ensures faster training and good generalization performance, making the GBU-TSVM suitable for  complex classification tasks. 
\end{itemize}

The rest of this paper is organized as follows: Section \ref{sec:related_work} presents an overview of related work, offering detailed discussions on SVM, GB-SVM, TSVM, and U-TSVM. In Section \ref{sec:proposed_model}, the proposed Granular-ball U-TSVM model is introduced, with an in-depth explanation of its theoretical foundation and algorithmic implementation. Section \ref{sec:experiments} presents the experimental analysis, showcasing the performance of GBU-TSVM on various benchmark datasets. In conclusion, Section \ref{sec:conclusion} of the manuscript provides a summary of the main results and suggests possible avenues for further research.

\section{Related Work} \label{sec:related_work}
In this section, we discuss the baseline classification models.
\subsection{Support Vector Machines}
Support Vector Machines (SVMs) represent a category of supervised learning techniques extensively applied in tasks involving classification and regression. 
Introduced by Vapnik et al., SVMs have evolved into one of the most widely adopted and effective methodologies in the field of machine learning due to their robustness and effectiveness in handling the complex data. SVMs operate by finding the optimal separating hyperplane that maximizes the margin between the classes in the training data. The margin is defined as the distance between the hyperplane and the nearest data points from either class, known as support vectors. The intuition behind this approach is that maximizing the margin leads to better generalization performance on unseen data, reducing the risk of overfitting.\\











\textbf{Primal Formulation}\\
The primal optimization problem for SVM can be stated as follows:

\begin{equation}
\begin{aligned}
\label{eqn:1}
& \underset{\omega, b, \xi}{\text{min}}
& & \frac{1}{2} \|\omega\|^2 + C \sum_{i=1}^{k} \xi_i \\
& \text{subject to}
& & y_i(\omega^T x_i + b) \geq 1 - \xi_i, \quad \xi_i \geq 0, \quad i = 1, \dots, k,
\end{aligned}
\end{equation}
In this context, $C$ represents for the penalty parameter, while $\xi_i$ signifies the slack variables. The aim is to identify the optimal separating hyperplane $(\omega, b)$ in such that:

\begin{equation}
\omega^T x + b = 0.
\end{equation}

\vspace{0.3cm}
\textbf{Dual Formulation}

The primal problem can be transformed into its dual form, which is usually easier to handle computationally. The Wolfe Dual of the SVM problem is given by:

\begin{equation}
\begin{aligned}
& \underset{\alpha}{\text{max}}
& & \sum_{i=1}^{k} \alpha_i - \frac{1}{2} \sum_{i=1}^{k} \sum_{j=1}^{k} y_i y_j \alpha_i \alpha_j \langle x_i, x_j \rangle \\
& \text{subject to}
& & \sum_{i=1}^{k} y_i \alpha_i = 0, \quad 0 \leq \alpha_i \leq C, \quad i = 1, \dots, k,
\end{aligned}
\\
\end{equation}
Here, $\alpha_i$ represents the Lagrangian multipliers. The solution $\alpha^*$ yields the optimal separating hyperplane $(\omega, b)$ in the following way:

\begin{align}
\omega &= \sum_{i=1}^{k} \alpha_i^* y_i x_i, \\
b &= \frac{1}{N_b} \left( y_i - \sum_{j=1}^{k} \alpha_j^* y_j \langle x_i, x_j \rangle \right),
\end{align}

where $N_b$ represents the count of support vectors satisfying $0 < \alpha_i^* < C$.

\vspace{0.3cm}
\textbf{Kernel Trick}

One of the most powerful features of SVMs is their ability to handle non-linearly separable data by using the kernel trick. A kernel function $K(x_i, x_j)$ computes the dot product of the data points in a higher-dimensional feature space without explicitly mapping the data points into that space. Commonly used kernel functions include the linear, polynomial, and radial basis function (RBF) kernels. The dual problem incorporating a kernel function becomes:

\begin{equation}
\begin{aligned}
& \underset{\alpha}{\text{max}}
& & \sum_{i=1}^{k} \alpha_i - \frac{1}{2} \sum_{i=1}^{k} \sum_{j=1}^{k} y_i y_j \alpha_i \alpha_j K(x_i, x_j) \\
& \text{subject to}
& & \sum_{i=1}^{k} y_i \alpha_i = 0, \quad 0 \leq \alpha_i \leq C, \quad i = 1, \dots, k,
\end{aligned}
\\
\end{equation}

SVMs excel in high-dimensional spaces, even when samples are scarce. It offers flexibility through the specification of different kernel functions for the decision function. It may struggle with noisy or outlier-laden data.

\subsection{Granular Ball Support Vector Machines (GB-SVM)}
Granular Ball Support Vector Machines (GB-SVM),  proposed by Xia et al.(2022) \cite{xia2022gbsvm}, extends the traditional SVM framework to handle uncertainties and improve robustness in classification tasks. The primary idea behind GB-SVM is to represent data points as granular balls rather than individual points. Each granular ball encapsulates a cluster of data points, providing a way to handle variability and uncertainty within the data. This representation leads to a more robust and efficient classification model, especially in scenarios where the data is noisy or contains outliers.

In GB-SVM, data points are grouped into granular balls, each characterized by a center and a radius. The center represents the mean of the data points within the ball, and the radius indicates the variability of the points around the center. This approach allows the model to account for the internal structure of the data clusters, leading to more accurate and stable classification boundaries.

\vspace{0.3cm}

The constraint for the support planes \( l'_1 \) and \( l'_2 \) is modified to incorporate the granular balls:
\begin{equation}
    y_i \omega \cdot c_i + y_i b - \|\omega\| r_i \geq 1
    \label{eq:constraint}
\end{equation}

The equations for the support planes \( l'_1 \) and \( l'_2 \) are:

\begin{equation}
\begin{aligned}
l'_1 &: \omega \cdot c_i - \|\omega\| r_i + b = 1, \quad y_i = +1 \\
l'_2 &: \omega \cdot c_i + \|\omega\| r_i + b = -1, \quad y_i = -1 \\
l_0' &: \omega \cdot c_i + b = 0
\end{aligned}
\end{equation}

The objective is to seek an optimal separation hyperplane \( (\omega, b) \) that accounts for the granular balls.\\

\textbf{Primal Formulation}
The primal optimization problem for an inseparable GB-SVM with slack variables \( \xi_i \) and a penalty coefficient \( C \) is formulated as follows:

\begin{equation}
\begin{aligned}
& \underset{\omega, b, \xi}{\text{min}}
& & \frac{1}{2} \|\omega\|^2 + C \sum_{i=1}^{k} \xi_i \\
& \text{subject to}
& & y_i (\omega \cdot c_i + b) - \|\omega\| r_i \geq 1 - \xi_i, \\
& & & \xi_i \geq 0, \quad i = 1, \dots, k,
\end{aligned}
\vspace{0.3cm}
\end{equation}

Here, \( \omega \) represents the weight vector, \( b \) is the bias term, \( \xi_i \) are the slack variables, \( c_i \) and \( r_i \) denote the center and radius of the \( i \)-th granular ball, respectively, and \( y_i \in \{+1, -1\} \) are the class labels. This formulation aims to find the optimal separating hyperplane that maximizes the margin while allowing for some misclassification controlled by the penalty parameter \( C \).\\

\textbf{Dual Formulation}

The Wolfe Dual of the GB-SVM problem introduces Lagrange multipliers \( \alpha_i \) and can be expressed as:

\begin{equation}
\begin{aligned}
& \underset{\alpha}{\text{max}}
& & -\frac{1}{2} \left( \sum_{i=1}^{k} \alpha_i y_i c_i \right)^2 - \frac{1}{2} \left( \sum_{i=1}^{k} \alpha_i r_i \right)^2 + \left\| \sum_{i=1}^{k} \alpha_i y_i c_i \right\| \sum_{i=1}^{k} \alpha_i r_i + \sum_{i=1}^{k} \alpha_i \\
& \text{subject to}
& &\quad \sum_{i=1}^{k} \alpha_i y_i = 0, \\
& & & 0 \leq \alpha_i \leq C, \quad i = 1, \dots, k,
\end{aligned}
\end{equation}

A simplified representation of the dual model (10) is:

\begin{equation}
\begin{aligned}
& \underset{\alpha}{\text{max}}
& & -\frac{1}{2} P^2 - \frac{1}{2} Q^2 + \|P\| Q + \sum_{i=1}^{k} \alpha_i \\
& \text{subject to}
& & \quad\sum_{i=1}^{k} \alpha_i y_i = 0, \\
& & & 0 \leq \alpha_i \leq C, \quad i = 1, \dots, k,
\end{aligned}
\vspace{0.3cm}
\end{equation}

where \( P = \sum_{i=1}^{k} \alpha_i y_i c_i \) and \( Q = \sum_{i=1}^{k} \alpha_i r_i \). The solution to this dual problem provides the necessary parameters to define the optimal separating hyperplane \( (\omega, b) \).\\

To find the optimal value of \( \alpha \), solve the convex optimization problem (11). Substituting these optimal \( \alpha \) values into the expressions for \( P \) and \( Q \) yields the most effective value for \( (\omega, b) \). \(\omega\)  can be given as:


\[
\omega = \frac{(\|P\| - Q)P}{\|P\|}
\]

where:

\[
\|\omega\| = \|P\| - Q
\]

After determining the most suitable \( \omega \), the intercept \( b \) can be computed using the support vectors. For each support vector \( i \) with \( \alpha_i \) neither 0 nor \( C \):\\
For positive support vectors (\( y_i = +1 \)):
\[
b = 1 - \omega \cdot c_i + \|\omega\| r_i
\]
For negative support vectors (\( y_i = -1 \)):
\[
b = -1 - \omega \cdot c_i - \|\omega\| r_i
\]
To obtain a stable \( b \), average the \( b \) values from all support vectors:

\[
b = \frac{1}{n_s} \sum_{i=1}^{n_s} \left( y_i - \omega \cdot c_i + y_i \|\omega\| r_i \right)
\]
Here, \( n_s \) denotes the count of support vectors, \( c_i \) represents the center of the \( i \)-th granular ball, and \( r_i \) signifies the radius of the \( i \)-th granular ball.

By representing data as granular balls, GB-SVM can better handle noise and outliers, leading to more stable classification boundaries. This sets the stage for further advancements, such as integrating Universum data to improve the model's contextual understanding and generalization capabilities, as explored in subsequent sections.

\subsection{Twin Support Vector Machine (TSVM)}
Twin Support Vector Machine (TSVM) is a binary classification algorithm introduced by Jayadeva et al. (2007) \cite{Jayadeva2007twin}. It differentiates itself from traditional SVMs by constructing two non-parallel hyperplanes instead of a single hyperplane. These two hyperplanes are strategically positioned such that one hyperplane is closer to one class, while the other hyperplane maintains a specific distance from the other class. This approach aims to enhance the classification performance, particularly for imbalanced datasets.

Let's examine two-class categorization scenario with $m_1$ Class 1(+ve) instances and $m_2$ Class 2(-ve) instances ($m_1 + m_2 = m$). The Class 1(+ve) data points are denoted by $A \in \mathbb{R}^{m_1 \times n}$, with each row $A_i \in \mathbb{R}^n$ representing a single data point. Similarly, the Class 2(-ve) data points are represented by $B \in \mathbb{R}^{m_2 \times n}$.

TSVM constructs two non-parallel hyperplanes:
\begin{equation}
f_+(x) = (\omega_+ \cdot x) + b_+ = 0 \quad \text{and} \quad f_-(x) = (\omega_- \cdot x) + b_- = 0,
\end{equation}

where $\omega_+, \omega_- \in \mathbb{R}^n$ and $b_+, b_- \in \mathbb{R}$. Each hyperplane is designed to be closer to one class and maintains a at least a distance of one from the other class.

\textbf{Primal Formulation:}
For determining the optimal hyperplanes, TSVM solves two QPPs:
\begin{equation}
\begin{aligned}
\min_{\omega_+, b_+, \xi} \quad & \frac{1}{2} \|A \omega_+ + e_+ b_+\|^2 + c_1 e_-^T \xi, \\
\text{subject to} \quad & - (B \omega_+ + e_- b_+) + \xi \geq e_-, &\xi \geq 0,
\end{aligned}
\\
\end{equation}

\begin{center}
and
\end{center}
\begin{equation}
\begin{aligned}
\min_{\omega_-, b_-, \eta} \quad & \frac{1}{2} \|B \omega_- + e_- b_-\|^2 + c_2 e_+^T \eta, \\
\text{subject to} \quad & (A \omega_- + e_+ b_-) + \eta \geq e_+, 
                       & \eta \geq 0.
\end{aligned}
\end{equation}
where $c_1, c_2 \geq 0$ are regularization parameters, and $e_+, e_-$ are vectors of ones of appropriate dimensions for positive and negative classes, respectively.\\

\textbf{Dual Formulation}
\vspace{0.3cm}

By introducing Lagrange multipliers $\alpha_i$ and $\beta_i$, we can derive the dual form of the QPPs (13) and (14):

\begin{equation}
\begin{aligned}
\max_{\alpha} \quad & e_-^T \alpha - \frac{1}{2} \alpha^T G (H^T H)^{-1} G^T \alpha, \\
\text{subject to} \quad & 0 \leq \alpha \leq c_1 e_-,
\end{aligned}
\label{eq:dual_problem_1}
\end{equation}

\begin{center}
    and
\end{center}
\begin{equation}
\begin{aligned}
\max_{\beta} \quad & e_+^T \beta - \frac{1}{2} \beta^T P (Q^T Q)^{-1} P^T \beta, \\
\text{subject to} \quad & 0 \leq \beta \leq c_2 e_+.
\end{aligned}
\label{eq:dual_problem_2}
\end{equation}

where  $H = [A \; e_+]$, $G = [B \; e_-]$, $P = [A \; e_+]$, and $Q = [B \; e_-]$.\\

The solutions $\alpha$ and $\beta$ from the dual problems provide the parameters for the hyperplanes as:

\begin{equation}
\begin{aligned}
v_1 &= -(H^T H)^{-1} G^T \alpha, \quad \text{where} \quad v_1 = \begin{bmatrix} \omega_+ \\ b_+ \end{bmatrix}, \\[6pt]
v_2 &= -(Q^T Q)^{-1} P^T \beta, \quad \text{where} \quad v_2 = \begin{bmatrix} \omega_- \\ b_- \end{bmatrix}.
\end{aligned}
\end{equation}
For non-linear classification tasks, these formulations can be extended using kernel functions, allowing the data to be mapped into a higher-dimensional feature space where the linear separation might be more feasible.

\subsection{Universum-Twin Support Vector Machine (U-TSVM)}
Universum-Twin Support Vector Machine (U-TSVM), introduced by Qi et al. (2012) \cite{qi2012twin}, incorporates supplementary data, known as Universum data, to enhance classification performance by leveraging information related to the underlying data distribution. This section elaborates on how U-TSVM utilizes Universum data to improve the robustness of the classification model.

In the U-TSVM framework, the training dataset $\tilde{D}$ includes both annotated data $D$ and Universum data $U$:

\begin{equation}
\tilde{D} = D \cup U,
\end{equation}
Here, the labeled/annotated data $D$ and the Universum data $U$ are specified as:
\begin{equation}
D = \{(z_1, \gamma_1), \ldots, (z_m, \gamma_m)\} \in (\mathbb{R}^n \times \Gamma)^m, \quad U = \{u_1, \ldots, u_p\} \in \mathbb{R}^n,
\end{equation}

with $z_i \in \mathbb{R}^n$, $\gamma_i \in \Gamma = \{-1, 1\}$ for $i = 1, \ldots, m$, and $u_j \in \mathbb{R}^n$ for $j = 1, \ldots, p$. The objective is to derive a function $\gamma = \text{sgn}(h(z))$ that can predict the label $\gamma$ for any given input datapoint $z \in \mathbb{R}^n$.\\

\textbf{Primal Formulation}

\vspace{0.3cm}
U-TSVM classifies data points into +1 (+ve Class) or -1 (-ve Class) by their distance to two nonparallel hyperplanes. The method incorporates Universum data by adding hinge loss functions to the quadratic programming problems:

\begin{align}
& \min_{\omega_+, b_+, \xi, \psi} \quad \quad \frac{1}{2} \|A \omega_+ + e_+ b_+\|^2 + c_1 e_-^T \xi + c_u e_u^T \psi, \\
& \text{subject to} \quad - (B \omega_+ + e_- b_+) + \xi \geq e_-, \quad \xi \geq 0, \\
& \quad \quad \quad  \quad \quad \quad (U \omega_+ + e_u b_+) + \psi \geq (-1 + \epsilon)e_u, \quad \psi \geq 0, \nonumber
\end{align}
\begin{center}
    and
\end{center}
\begin{align}
& \min_{\omega_-, b_-, \eta, \psi^*} \quad \quad \frac{1}{2} \|B \omega_- + e_- b_- \|^2 + c_2 e_+^T \eta + c_u e_u^T \psi^*, \\
& \text{subject to} \quad (A \omega_- + e_+ b_-) + \eta \geq e_+, \quad \eta \geq 0, \\
& \quad \quad \quad  \quad  \quad - (U \omega_- + e_u b_-) + \psi^* \geq (-1 + \epsilon)e_u, \quad \psi^* \geq 0. \nonumber
\end{align}

Here, $\psi = (\psi_1, \ldots, \psi_p)^T$, $\psi^* = (\psi^*_1, \ldots, \psi^*_p)^T$, and $\epsilon, c_1, c_2, c_u \in [0, \infty)$ are predefined parameters. Vectors $e_+, e_-, e_u$ represent ones of suitable dimensions, and $U \in \mathbb{R}^{p \times n}$ denotes the Universum class, with each row $U_i \in \mathbb{R}^n$ as a Universum sample.
The hinge loss terms for Universum samples are given by:
\begin{align}
\psi_i &= \max\{0, -1 + \epsilon - h_+(u_i)\}, \\
\psi^*_i &= \max\{0, -1 + \epsilon + h_-(u_i)\},
\end{align}
for $i = 1, \ldots, p$. The total hinge loss approximates the loss function in the objective.

We define matrices and vectors as follows:
\begin{equation}
H = [A \; e_+], \quad G = [B \; e_-], \quad O = [U \; e_u],
\end{equation}
and the extended vector $\vartheta_+ = [\omega_+ \; b_+]^T$ represents the parameters of the separating hyperplane for the positive class. This equality can be reformulated as follows:\\
\begin{equation}
H^T H \vartheta_+ + G^T \alpha - O^T \mu = 0,
\end{equation}
leading to:
\begin{equation}
\vartheta_+ = -(H^T H)^{-1} (G^T \alpha - O^T \mu).
\end{equation}

\textbf{Dual Formulation}

\vspace{0.3cm}
The dual formulations for the optimization problems involve Lagrange multipliers $\alpha$ and $\mu$ for the positive hyperplane:
\begin{align}
\max_{\alpha, \mu} & \quad - \frac{1}{2} (\alpha^T G - \mu^T O) (H^T H)^{-1} (G^T \alpha - O^T \mu) \\
& \quad + e_-^T \alpha + (\epsilon - 1)e_u^T \mu, \nonumber \\
\text{subject to} & \quad 0 \leq \alpha \leq c_1 e_-, \nonumber \\
& \quad 0 \leq \mu \leq c_u e_u, \nonumber
\end{align}
and for the negative hyperplane:
\begin{align}
\max_{\lambda, \nu} & \quad - \frac{1}{2} (\lambda^T P - \nu^T S) (Q^T Q)^{-1} (P^T \lambda - S^T \nu) \\
& \quad + e_+^T \lambda + (\epsilon - 1)e_u^T \nu, \nonumber \\
\text{subject to} & \quad 0 \leq \lambda \leq c_2 e_+, \nonumber \\
& \quad 0 \leq \nu \leq c_u e_u, \nonumber
\end{align}
where
\begin{equation}
P = [A \; e_+], \quad Q = [B \; e_-], \quad S = [U \; e_u],
\end{equation}
and the extended vector $\vartheta_- = [\omega_- \; b_-]^T$ represents the parameters of the separating hyperplane for the negative class. This equality can be reformulated as follows:\\
\begin{equation}
Q^T Q \vartheta_- + P^T \lambda - S^T \nu = 0,
\end{equation}
thus,
\begin{equation}
\vartheta_- = -(Q^T Q)^{-1} (P^T \lambda - S^T \nu).
\end{equation}
The separating hyperplanes are derived from $\vartheta_+$ and $\vartheta_-$:
\begin{equation}
\omega_+^T z + b_+ = 0, \quad \omega_-^T z +  b_-- = 0.
\end{equation}
A distinct sample  $z \in \mathbb{R}^n$ is classified based on its minimum distance to these hyperplanes:
\begin{equation}
h(z) = \min_{+, -} \{ \delta_+(z), \delta_-(z) \},
\end{equation}
where
\begin{equation}
\delta_+(z) = |\omega_+^T z +  b_+|, \quad \delta_-(z) = |\omega_-^T z +  b_-|.
\end{equation}

\section{Proposed granular ball \label{sec:proposed_model}twins support vector machine with universum data (GBU-TSVM) Model}
In this section, we discuss the formulation of the proposed granular ball twins support vector machine with universum data for linear and non-linear cases.

\subsection{Linear Case}
In the foundational work on classification and clustering utilizing granular-ball computing, a granular-ball \(GB_j\) is defined as \(GB_j = \{z_i \mid i = 1, 2, \ldots, k\}\). The center of a granular-ball, denoted as \(c\), is computed using the formula \(c = \frac{1}{k} \sum_{i=1}^{k} z_i\), where \(z_i\) represents an individual sample within the granular-ball, and \(k\) is the total number of samples in the granular-ball. The radius of a granular-ball can be determined through several methods, with the two primary methods being the average distance and the maximum distance. The average distance is calculated as \(r_{\text{avg}} = \frac{1}{k} \sum_{i=1}^{k} \|z_i - c\|\), while the maximum distance is given by \(r_{\text{max}} = \max \|z_i - c\|\), where \(\|\cdot\|\) represents the Euclidean distance.\\

Given a dataset $\mathcal{D}$ with $n$ samples, $z_i \in \mathcal{D}$. $GB_j$ ($j = 1, 2, \ldots, m$) represents a granular-ball generated from dataset $\mathcal{D}$, with $m$ being the total number of granular-balls. The optimization model is structured as follows:

\begin{align}
&\hspace{1cm}f(z_i, \vec{\alpha}) \rightarrow g(GB_j, \vec{\beta}), \\
&\text{subject to} \quad \min \sum_{j=1}^{m} |GB_j| + m + \text{loss}(GB_j), 
\quad \text{quality}(GB_j) \geq T,
\end{align}

where $|\cdot|$ denotes cardinality, $f$ represents traditional learning methods, $g$ denotes granular-ball learning methods, and $\text{loss}(GB_j)$ optimizes granular-ball quality.
Expanding from the basics of granular-ball computing, we now integrate Universum data to enrich our classification approach. 
The classification problem with Universum is represented as follows:\\

Let $\tilde{\mathcal{T}}$ denote the augmented training set comprising labeled granular-balls and Universum samples:
\begin{equation}
\tilde{\mathcal{T}} = \mathcal{T}_+ \cup \mathcal{T}_- \cup \mathcal{U},
\end{equation}
where $\cup$ denotes set union. The components of $\tilde{\mathcal{T}}$ are defined as follows:
\begin{equation}
\mathcal{T}_+ = \{(c_1^+, r_1^+, +1), \ldots, (c_{m_1}^+, r_{m_1}^+, +1)\} \subset (\mathbb{R}^n \times \mathbb{R}_{\geq 0} \times \mathcal{Y})^{m_1},
\end{equation}
\begin{equation}
\mathcal{T}_- = \{(c_1^-, r_1^-, -1), \ldots, (c_{m_2}^-, r_{m_2}^-, -1)\} \subset (\mathbb{R}^n \times \mathbb{R}_{\geq 0} \times \mathcal{Y})^{m_2},
\end{equation}
\begin{equation}
\mathcal{U} = \{c_1^*, \ldots, c_u^*\} \subset (\mathbb{R}^n)^u,
\end{equation}
\begin{equation}
\mathcal{R}_u = \{r_1^*, \ldots, r_u^*\} \subset (\mathbb{R})^u
\end{equation}
with:
\renewcommand{\labelitemi}{$\diamond$}
\begin{itemize}
\item $c_i^+ \in \mathbb{R}^n, r_i^+ \in \mathbb{R}_{\geq 0}, i = 1, \ldots, m_1$: centroids and radii for positive (+1) class granular-balls.
\item $c_j^- \in \mathbb{R}^n, r_j^- \in \mathbb{R}_{\geq 0}, j = 1, \ldots, m_2$: centroids and radii for negative (-1) class granular-balls.
\item $c_k^* \in \mathbb{R}^n, r_k^* \in \mathbb{R}_{\geq 0}, k = 1, \ldots, u$: centroids and radii for Universum samples.
\item $\mathcal{Y} = \{-1, +1\}$: binary class labels.
\end{itemize}
Here, $\mathcal{T}_+$ and $\mathcal{T}_-$ represent the labeled training data in the form of granular-balls. Each granular-ball is characterized by its centroid $c_i^+$ or $c_j^-$, radius $r_i^+$ or $r_j^-$, and class label $+1$ or $-1$, respectively. The set $\mathcal{U}$ represents the Universum data, consisting of unlabeled samples that do not belong to either class but provide valuable domain information. $\mathcal{R}_u$ represents the radii of the granular-balls generated from the Universum data.\\

The parameters $\mathcal{A}$, $\mathcal{B}$, $\mathcal{R}_+$, $\mathcal{R}_-$, and $\mathcal{R}_u$ are defined as follows:
\renewcommand{\labelitemi}{$\diamond$}
\begin{itemize}
\item $\mathcal{A} = \{c_1^+, \ldots, c_{m_1}^+\}$ denotes the centroids of granular-balls in the positive (+1) class.
\item $\mathcal{B} = \{c_1^-, \ldots, c_{m_2}^-\}$ denotes the centroids of granular-balls in the negative (-1) class.
\item $\mathcal{R}_+ = \{r_1^+, \ldots, r_{m_1}^+\}$ represents the radii associated with the centroids in $\mathcal{A}$.
\item $\mathcal{R}_- = \{r_1^-, \ldots, r_{m_2}^-\}$ represents the radii associated with the centroids in $\mathcal{B}$.
\end{itemize}
The primary objective of GBU-TSVM is to derive a decision function expressed as:
\begin{equation}
\gamma = \text{sgn}(h(z)),
\end{equation}
to predict the classification $\gamma \in \mathcal{Y}$ for any input data point $z \in \mathbb{R}^n$. Unlike traditional TSVMs and its variants, the function $h(z)$ in GBU-TSVM considers the distances from $z$ to the granular-balls of both classes, not just the distances to individual points or hyperplanes.\\

The $\epsilon$-insensitive loss function used in U-SVM is defined as follows:
\begin{equation}
\frac{1}{2} \|\omega\|_2^2 + c \sum_{i=1}^{l} \phi_{\epsilon}[\gamma_i f_{w,b}(z_i)] + d \sum_{j=1}^{u} \rho[f_{w,b}(z^*_j)],
\end{equation}
Here, $\max\{0, \epsilon - \gamma_i f(z_i)\}$ denotes the hinge loss function (\(\phi_{\epsilon}\)[t]), integrating prior information conveyed by the Universum. The function \(\rho[t] = \rho_{-\epsilon}[t] + \rho_{-\epsilon}[-t]\) accounts for error tolerance within a specified margin. The sum of the losses captures the prior knowledge incorporated from the Universum :
$\sum_{j=1}^{u} \rho[f_{w,b}(z^*_j)]$. A smaller value of this sum suggests a higher probability for the classifier \(f_{w,b}\), and vice versa.\\

In U-TSVM, the hinge loss function is modified to include prior knowledge from Universum samples, making it approximately equal to the sum of $\psi_i$ and $\psi_i^*$. These terms are defined as:

\begin{equation}
\psi_i = \begin{cases}
\quad\quad\quad\quad0 \;, & f_{\omega_+, b_+}(z^*_i) \geq -1 + \epsilon \\
-1 + \epsilon - f_{\omega_+, b_+}(z^*_i), & f_{\omega_+, b_+}(z^*_i) < -1 + \epsilon
\end{cases}
\end{equation}
\begin{equation}
= \max\{0, -1 + \epsilon - f_{\omega_+, b_+}(z_i^*)\},
\end{equation}
\begin{center}
    and
\end{center}

\begin{equation}
\psi^*_i = \begin{cases}
\quad\quad\quad\quad0 \;, & -f_{\omega_-, b_-}(z^*_i) \geq -1 + \epsilon \\
-1 + \epsilon + f_{\omega_-, b_-}(z^*_i), & -f_{\omega_-, b_-}(z^*_i) < -1 + \epsilon
\end{cases}
\end{equation}
\begin{equation}
= \max\{0, -1 + \epsilon - f_{\omega_-, b_-}(z_i^*) \},
\end{equation}
where $i = 1, \ldots, u$. These functions adjust the loss based on the distance of Universum samples from the decision boundary.\\
\textbf{Primal Formulation}

\vspace{0.3cm}
For GBU-TSVM, the hinge loss function considers the centroids and radii of Universum samples. The modified loss functions are given by:
\begin{equation}
\psi_i = \begin{cases}
\quad\quad\quad\quad0 \;, & f_{\omega_+, b_+}(c_i^*) + r_i^* \geq -1 + \epsilon \\
-1 + \epsilon - f_{\omega_+, b_+}(c_i^*) - r_i^*\;, & f_{\omega_+, b_+}(c_i^*) + r_i^* < -1 + \epsilon
\end{cases}
\end{equation}
\begin{equation}
= \max\{0, -1 + \epsilon - f_{\omega_+, b_+}(c_i^*) - r_i^*\},
\end{equation}
\begin{center}
    and
\end{center}
\begin{equation}
\psi_i^* = \begin{cases}
\quad\quad\quad\quad0 \;, & -f_{\omega_-, b_-}(c_i^*) + r_i^* \geq -1 + \epsilon \\
-1 + \epsilon + f_{\omega_-, b_-}(c_i^*) - r_i^* \;,  & -f_{\omega_-, b_-}(c_i^*) + r_i^* < -1 + \epsilon
\end{cases}
\end{equation}
\begin{equation}
= \max\{0, -1 + \epsilon + f_{\omega_-, b_-}(c_i^*) - r_i^* \}.
\end{equation}
where $i = 1, \ldots, u$. These adjustments in the loss function allow GBU-TSVM to incorporate the geometry of Universum samples into the learning process.\\

For GBU-TSVM, a new granular ball point is classified as either +1 or -1 depending on its distance from the two nonparallel hyperplanes. The pre-existing knowledge contained in the Universum is incorporated by introducing modified hinge loss functions into the following quadratic programming problems (QPPs) respectively:

\begin{align}
& \min_{\omega_+, b_+, \xi, \psi} \quad \frac{1}{2} \|\mathcal{A} \omega_+ + e_+ b_+\|^2 + c_1 e_-^T \xi + c_u e_u^T \psi, \\
& \text{subject to} \quad - (\mathcal{B} \omega_+ + e_- b_+) + \xi \geq e_- - \mathcal{R}_-, \quad \xi \geq 0, \\
& \quad \quad \quad  \quad \quad \quad (\mathcal{U} \omega_+ + e_u b_+) + \psi \geq (-1 + \epsilon)e_u - \mathcal{R}_u
, \quad \psi \geq 0, \nonumber
\end{align}
\begin{center}
    and
\end{center}
\begin{align}
& \min_{\omega_-, b_-, \eta, \psi^*} \quad \frac{1}{2} \|\mathcal{B} \omega_- + e_- b_- \|^2 + c_2 e_+^T \eta + c_u e_u^T \psi^*, \\
& \text{subject to} \quad (\mathcal{A} \omega_- + e_+ b_-) + \eta \geq e_+ - \mathcal{R}_+, \quad \eta \geq 0, \\
& \quad \quad \quad  \quad  \quad - (\mathcal{U} \omega_- + e_u b_-) + \psi^* \geq (-1 + \epsilon)e_u - \mathcal{R}_u, \quad \psi^* \geq 0, \nonumber
\end{align}
where $\psi = (\psi_1, \ldots, \psi_u)^T$ and $\psi^* = (\psi^*_1, \ldots, \psi^*_u)^T$. The constants $\epsilon$, $c_1$, $c_2$, and $c_u$ are all non-negative, i.e., $\epsilon, c_1, c_2, c_u \geq 0$. The vectors $e_+$, $e_-$, and $e_u$ are unit vectors of dimensions $m1$, $m2$, and $u$, respectively. The Universum class, $\mathcal{U} \in \mathbb{R}^{u \times n}$, consists of rows $\mathcal{U}_i \in \mathbb{R}^n$, each representing a Universum sample. $\mathcal{R}_u$ denotes the radii of the granular balls generated from the Universum data.\\

\begin{figure}[h]
  \centering
  \includegraphics[width=1.0\linewidth]{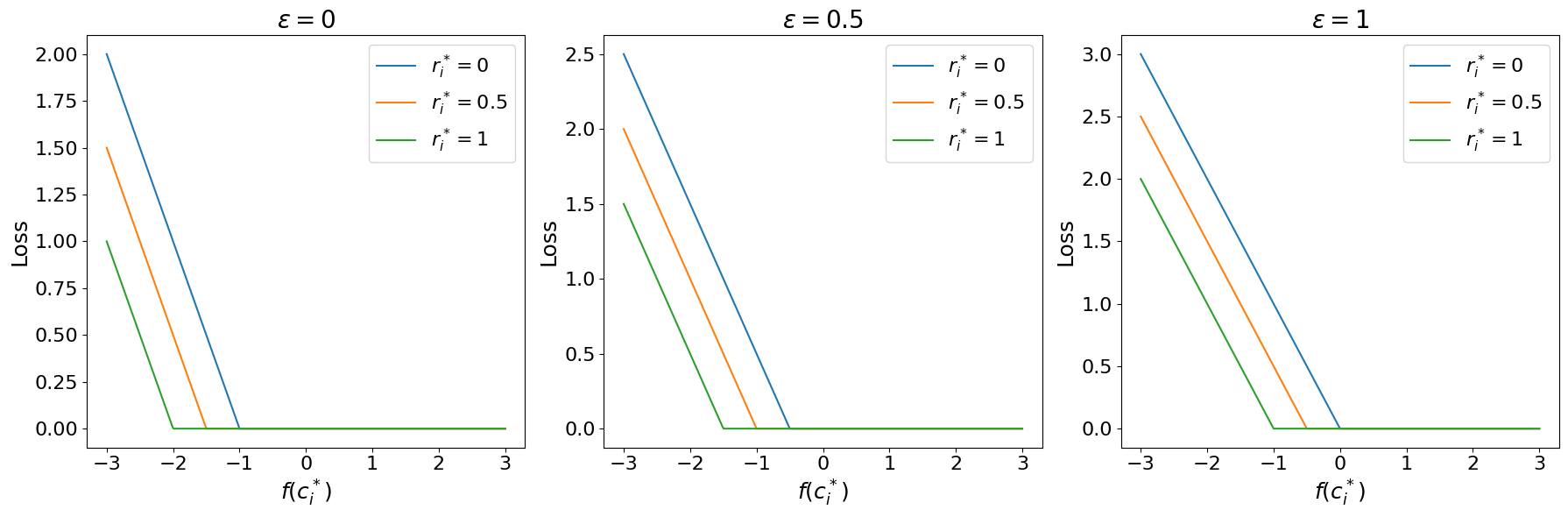}
  \caption{Hinge Loss $\psi$}
  \label{fig:hinge1}
\end{figure}

\begin{figure}[h]
  \centering
  \includegraphics[width=1.0\linewidth]{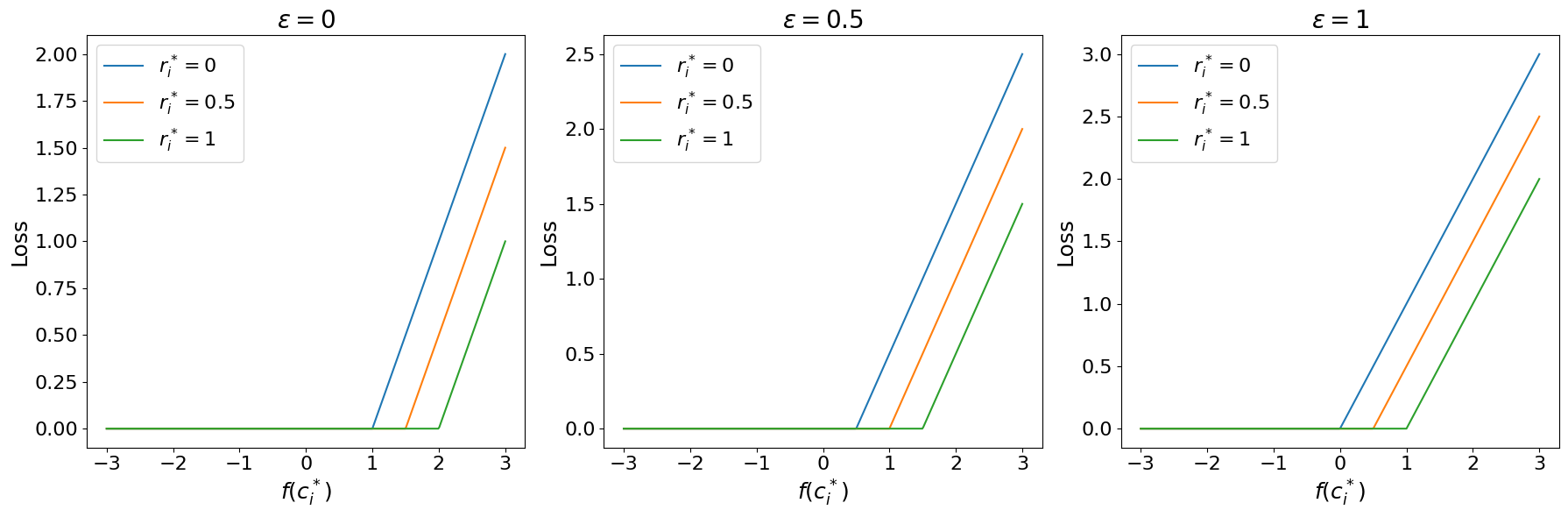}
  \caption{Hinge Loss $\psi^*$}
  \label{fig:hinge2}
\end{figure}
The Lagrangian associated with equations (54)–(55) can be expressed as:
\begin{align}
L(\Omega) &= \frac{1}{2} \|\mathcal{A} \omega_+ + e_+ b_+\|^2 + c_1 e_-^T \xi + c_u e_u^T \psi \nonumber\\
&\quad - \alpha^T (- (\mathcal{B} \omega_+ + e_- b_+) + \xi - e_- + \mathcal{R}_-) \nonumber\\
&\quad - \beta^T \xi - \mu^T ((\mathcal{U} \omega_+ + e_u b_+) + \psi + (1 - \epsilon)e_u + \mathcal{R}_u) - \gamma^T \psi,
\end{align}
where $\Omega = \{\omega_+, b_+, \xi, \psi, \alpha, \beta, \mu, \gamma\}$. The Lagrange multipliers are represented by the vectors $\alpha = (\alpha_1, \ldots, \alpha_{m_1})^T$, $\beta = (\beta_1, \ldots, \beta_{m_1})^T$, $\mu = (\mu_1, \ldots, \mu_u)^T$, and $\gamma = (\gamma_1, \ldots, \gamma_u)^T$.\\




\textbf{Dual Formulation}

\vspace{0.3cm}
Using the Karush-Kuhn-Tucker (KKT) conditions, we have:
\begin{align}
& \max_{\Omega} \quad L(\Omega), \\
& \text{subject to} \quad \frac{\partial L(\Omega)}{\partial \omega_+} = \mathbf{0}, \quad \frac{\partial L(\Omega)}{\partial b_+} = 0, \quad \frac{\partial L(\Omega)}{\partial \xi} = \mathbf{0}, \quad \frac{\partial L(\Omega)}{\partial \psi} = \mathbf{0}, \\
& \quad \alpha, \beta, \mu, \gamma \geq 0.
\end{align}

Following (60), the outcomes are as follows:
\begin{equation}
\frac{\partial L}{\partial \omega_+} = \mathcal{A}^T (\mathcal{A} \omega_+ + e_+ b_+) + \mathcal{B}^T \alpha - \mathcal{U}^T \mu = 0,
\end{equation}

\begin{equation}
\frac{\partial L}{\partial b_+} = e_+^T (\mathcal{A} \omega_+ + e_+ b_+) + e^T \alpha - e^T \mu = 0,
\end{equation}

\begin{equation}
\frac{\partial L}{\partial \xi} = c_1 e_- - \alpha - \beta = 0,
\end{equation}

\begin{equation}
\frac{\partial L}{\partial \psi} = c_u e_u - \mu - \gamma = 0.
\end{equation}

Given that \(\beta, \gamma \geq 0\), equations (64) - (65) become:

\begin{equation}
0 \leq \alpha \leq c_1 e_-,
\end{equation}

\begin{equation}
0 \leq \mu \leq c_u e_u.
\end{equation}
From (62) and (63), we have:
\begin{equation}
[\mathcal{A}\;e_+]^T [\mathcal{A}\;e_+] [\omega_+\;b_+]^T + [\mathcal{B}\;e_-]^T \alpha - [\mathcal{U} \; e_u]^T \mu = 0.
\end{equation}
We define matrices and vectors as follows:
\begin{equation}
H = [\mathcal{A}\;e_+],\quad G = [\mathcal{B}\;e_-],\quad O = [\mathcal{U}\;e_u],
\end{equation}
and the extended vector  $\vartheta_+ = [\omega_+\;b_+]^T$ represents the parameters of the separating
hyperplane for the positive class. Eq.(68) can be reformulated as follows:\\
\begin{equation}
H^T H \vartheta_+ + G^T \alpha - O^T \mu = 0,
\end{equation}
leading to:
\begin{equation}
\vartheta_+ = -(H^T H)^{-1} (G^T \alpha - O^T \mu).
\end{equation}

In dual optimization theory, the Wolfe dual corresponding to the (54)–(55) is formulated as

\begin{align}
&\begin{aligned}
\max_{\alpha, \mu} & \quad \quad- \frac{1}{2} (\alpha^T G - \mu^T O) (H^T H)^{-1} (G^T \alpha - O^T \mu)  \\
& \quad \quad + (e_-^T - \mathcal{R}_- ^T) \alpha \; + \;((\epsilon - 1)e_u^T - \mathcal{R}_u^T) \mu,
\end{aligned} \\
&\begin{aligned}
\text{subject to} & \quad 0 \leq \alpha \leq c_1 e_-, \\
& \quad 0 \leq \mu \leq c_u e_u,
\end{aligned}
\end{align}

Analogously, the dual form for (56)–(57) and the optimization problem corresponding to generation of negative hyperplane is as follows:

\begin{align}
&\begin{aligned}
\max_{\lambda, \nu} & \quad \quad - \frac{1}{2} (\lambda^T P - \nu^T S) (Q^T Q)^{-1} (P^T \lambda - S^T \nu) \\
& \quad \quad + (e_+^T - \mathcal{R}_+ ^T) \lambda \; + \; ((\epsilon - 1)e_u^T - \mathcal{R}_u^T) \nu,
\end{aligned} \\
&\begin{aligned}
\text{subject to} & \quad 0 \leq \lambda \leq c_2 e_+, \\
& \quad 0 \leq \nu \leq c_u e_u,
\end{aligned}
\end{align}
where

\begin{equation}
P = [\mathcal{A}\;e_+],\quad Q = [\mathcal{B}\;e_-],\quad S = [\mathcal{U}\;e_u],
\end{equation}

and the extended vector $\vartheta_- = [\omega_- \; b_-]^T$, represents the parameters of the separating hyperplane for the negative class. This equality can be reformulated as follows:\\
\begin{equation}
Q^T Q \vartheta_- + P^T \lambda - S^T \nu = 0,
\end{equation}
thus,

\begin{equation}
\vartheta_- = -(Q^T Q)^{-1} (P^T \lambda - S^T \nu).
\end{equation}
The separating hyperplanes are derived from $\vartheta_+$ and $\vartheta_-$ Eq.(71) and (78),:
\begin{equation}
\omega_+^T z + b_+ = 0, \quad \omega_-^T z +  b_-- = 0.
\end{equation}
A distinct sample  $z \in \mathbb{R}^n$ is classified based on its minimum distance to these hyperplanes:
\begin{equation}
h(z) = \min_{+, -} \{ \delta_+(z), \delta_-(z) \},
\end{equation}
Here,
\begin{equation}
\delta_+(z) = |\omega_+^T z +  b_+|, \quad \delta_-(z) = |\omega_-^T z +  b_-|.
\end{equation}

where \( |\cdot| \) signifies the orthogonal distance of the point \( z \) from the planes $\omega_+^T z + b_+$ and $\omega_-^T z + b_-$

\begin{figure}[h]
  \centering
  \includegraphics[width=0.6\linewidth]{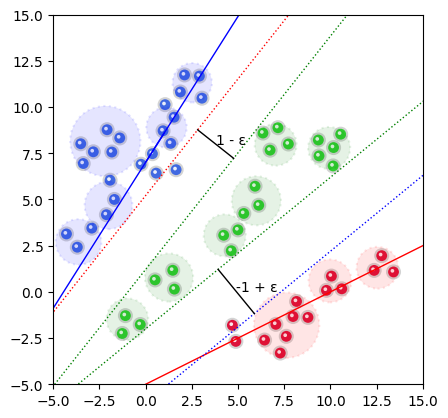}
  \caption{GBU-TSVM Geometric Interpretation}
  \label{fig:GBUTSVMgeometric}
\end{figure}

\subsection{Non-linear Case for GBU-TSVM using RBF Kernel}
\vspace{0.3cm}
To extend the Granular Ball TSVM with Universum data (GBU-TSVM) to the non-linear case, we utilize the kernel trick. Specifically, we employ the Radial Basis Function (RBF) kernel. The general approach is to map the data points into a higher-dimensional feature space, perform the required computations, and then derive the decision functions. Generating granular-balls in high-dimensional space is challenging because the radius cannot be directly mapped. Mapping original space balls to kernel space changes the data distribution, making them inaccurate. The best solution is to generate granular-balls directly in the kernel space.\\

Let $\phi: \mathbb{R}^n \rightarrow \mathcal{H}$ denote the mapping from the input space to the feature space induced by the RBF kernel. The RBF kernel is defined as:
\begin{equation}
K<z_i, z_j> = \exp\left(-\frac{\|z_i - z_j\|^2}{2\sigma^2}\right)
\end{equation}
where $\sigma$ is a hyperparameter controlling the width of the Gaussian function.
Given a dataset $D$ with $n$ samples, $x_i \in D$, we first map all the data points to the feature space:
\begin{equation}
\phi(z_i), \quad i = 1, 2, \ldots, n
\end{equation}
We form granular balls in the feature space. For a granular ball $GB_j$ with $k$ samples, the center and radius in the feature space are defined as:
\begin{equation}
c_j = \frac{1}{k} \sum_{i=1}^{k} \phi(z_i)
\end{equation}
\begin{equation}
r_j = \frac{1}{k} \sum_{i=1}^{k} \|\phi(z_i) - c_j\|
\end{equation}

Let $\tilde{\mathcal{T}}$ denote the augmented training set comprising labeled granular-balls and Universum samples:
\begin{equation}
\tilde{\mathcal{T}} = \mathcal{T}_+ \cup \mathcal{T}_- \cup \mathcal{U},
\end{equation}
where $\cup$ denotes set union. The components of $\tilde{\mathcal{T}}$ are defined as follows:
\begin{equation}
\mathcal{T}_+ = \{(c_1^+, r_1^+, +1), \ldots, (c_{m_1}^+, r_{m_1}^+, +1)\} \subset (\mathbb{R}^n \times \mathbb{R}_{\geq 0} \times \mathcal{Y})^{m_1},
\end{equation}
\begin{equation}
\mathcal{T}_- = \{(c_1^-, r_1^-, -1), \ldots, (c_{m_2}^-, r_{m_2}^-, -1)\} \subset (\mathbb{R}^n \times \mathbb{R}_{\geq 0} \times \mathcal{Y})^{m_2},
\end{equation}
\begin{equation}
\mathcal{U} = \{c_1^*, \ldots, c_u^*\} \subset (\mathbb{R}^n)^u,
\end{equation}
\begin{equation}
\mathcal{R}_u = \{r_1^*, \ldots, r_u^*\} \subset (\mathbb{R})^u
\end{equation}
We examine the following hyperplanes generated by the kernel:
\begin{align}
\mathcal{K}(z,\mathcal{C^T}) \omega_+ + b_+ &= 0, \\
\mathcal{K}(z, \mathcal{C^T}) \omega_-   + b_- &= 0,
\end{align}
where $\mathcal{C^T} = [\mathcal{A} \; \mathcal{B}]^T$ and $\mathcal{K}$ is the chosen Gaussian kernel function.\\
\textbf{Primal Formulation}
\vspace{0.3cm}
The nonlinear optimization problems can be stated as:
\begin{align}
& \min_{\omega_+, b_+, \xi, \psi} \quad \frac{1}{2} \|\mathcal{K}(\mathcal{A}, \mathcal{C^T}) \omega_+ + e_+ b_+\|^2 + c_1 e_-^T \xi + c_u e_u^T \psi, \\
& \text{subject to} \quad - (\mathcal{K}(\mathcal{B}, \mathcal{C^T}) \omega_+ + e_- b_+) + \xi \geq e_- - \mathcal{R}_-, \quad \xi \geq 0, \\
& \quad \quad \quad  \quad \quad \quad (\mathcal{K}(\mathcal{U}, \mathcal{C^T}) \omega_+ + e_u b_+) + \psi \geq (-1 + \epsilon)e_u - \mathcal{R}_u, \quad \psi \geq 0, \nonumber
\end{align}
\begin{center}
    and
\end{center}
\begin{align}
& \min_{\omega_-, b_-, \eta, \psi^*} \quad \frac{1}{2} \|\mathcal{K}(\mathcal{B}, \mathcal{C^T}) \omega_- + e_- b_- \|^2 + c_2 e_+^T \eta + c_u e_u^T \psi^*, \\
& \text{subject to} \quad (\mathcal{K}(\mathcal{A}, \mathcal{C^T}) \omega_- + e_+ b_-) + \eta \geq e_+ - \mathcal{R}_+, \quad \eta \geq 0, \\
& \quad \quad \quad  \quad  \quad - (\mathcal{K}(\mathcal{U}, \mathcal{C^T}) \omega_- + e_u b_-) + \psi^* \geq (-1 + \epsilon)e_u - \mathcal{R}_u, \quad \psi^* \geq 0, \nonumber
\end{align}

\vspace{0.3cm}

\textbf{Dual Formulation}

\vspace{0.3cm}
Following the same procedure as in the linear case, we obtain the Wolfe dual of the first problem (93)-(94):

\begin{align}
&\begin{aligned}
\max_{\alpha, \mu} & \quad \quad- \frac{1}{2} (\alpha^T G_{\Phi} - \mu^T O_{\Phi}) (H_{\Phi}^T H_{\Phi})^{-1} (G_{\Phi}^T \alpha - O_{\Phi}^T \mu)  \\
& \quad \quad + (e_-^T - \mathcal{R}_- ^T) \alpha \; + \;((\epsilon - 1)e_u^T - \mathcal{R}_u^T) \mu,
\end{aligned} \\
&\begin{aligned}
\text{subject to} & \quad 0 \leq \alpha \leq c_1 e_-, \\
& \quad 0 \leq \mu \leq c_u e_u,
\end{aligned}
\end{align}
We define matrices and vectors as follows:
\begin{equation}
H_{\Phi} = [\mathcal{K}(\mathcal{A}, \mathcal{C^T})\quad e_+],\quad G_{\Phi} = [\mathcal{K}(\mathcal{B}, \mathcal{C^T})\quad e_-],\quad O_{\Phi} = [\mathcal{K}(\mathcal{U}, \mathcal{C^T})\quad e_u],
\end{equation}
and the extended vector  $\vartheta_+ = [\omega_+\;b_+]^T$ represents the parameters of the separating
hyperplane for the positive class. This equality can be reformulated as follows:\\
\begin{equation}
H^T H \vartheta_+ + G^T \alpha - O^T \mu = 0,
\end{equation}
leading to:
\begin{equation}
\vartheta_+ = -(H_{\Phi}^T H_{\Phi})^{-1} (G_{\Phi}^T \alpha - O_{\Phi}^T \mu).
\end{equation}
Analogously, the dual form for (95)–(96) and the optimization problem corresponding to the generation of negative hyperplane is as follows:

\begin{align}
&\begin{aligned}
\max_{\lambda, \nu} & \quad \quad - \frac{1}{2} (\lambda^T P_{\Phi} - \nu^T S_{\Phi}) (Q_{\Phi}^T Q_{\Phi})^{-1} (P_{\Phi}^T \lambda - S_{\Phi}^T \nu) \\
& \quad \quad + (e_+^T - \mathcal{R}_+ ^T) \lambda \; + \; ((\epsilon - 1)e_u^T - \mathcal{R}_u^T) \nu,
\end{aligned} \\
&\begin{aligned}
\text{subject to} & \quad 0 \leq \lambda \leq c_2 e_+, \\
& \quad 0 \leq \nu \leq c_u e_u,
\end{aligned}
\end{align}
where

\begin{equation}
P_{\Phi} = [\mathcal{K}(\mathcal{A}, \mathcal{C^T})\quad e_+],\quad Q_{\Phi} = [\mathcal{K}(\mathcal{B}, \mathcal{C^T})\quad e_-],\quad S_{\Phi} = [\mathcal{K}(\mathcal{U}, \mathcal{C^T})\quad e_u],
\end{equation}

and the extended vector $\vartheta_- = [\omega_- \; b_-]^T$, represents the parameters of the separating hyperplane for the negative class. This equality can be reformulated as follows:\\
\begin{equation}
Q^T Q \vartheta_- + P^T \lambda - S^T \nu = 0,
\end{equation}
thus,

\begin{equation}
\vartheta_- = -(Q^T Q)^{-1} (P^T \lambda - S^T \nu).
\end{equation}
The separating hyperplanes are derived from $\vartheta_+$ and $\vartheta_-$ Eq.(101) and (106),:

\begin{align}
\mathcal{K}(z,\mathcal{C^T}) \omega_+ + b_+ &= 0, \quad
\mathcal{K}(z, \mathcal{C^T}) \omega_-   + b_- = 0,
\end{align}

A distinct sample  $z \in \mathbb{R}^n$ is classified based on its minimum distance to these hyperplanes:
\begin{equation}
h(z) = \min_{+, -} \{ \delta_+(z), \delta_-(z) \},
\end{equation}
Here,
\begin{equation}
\delta_+(z) = |\mathcal{K}(z,\mathcal{C^T}) \omega_+ + b_+|, \quad \delta_-(z) = |\mathcal{K}(z,\mathcal{C^T}) \omega_- + b_-|.
\end{equation}

and \( |\cdot| \) signifies the orthogonal distance of the point \( z \) from the planes generated by the kernel.

\section{Experiments} \label{sec:experiments}
In this section, we discuss about the datasets used for the evaluation of baseline and proposed model. Moreover, we evaluate the models statistically to evaluate their significance.
\subsection{Dataset Information}
Several binary class UCI benchmark datasets with binary class were thoughtfully selected for our experiments. The dataset details are summarized in Table \ref{tab:dataset_information}. 

\begin{table}[h]
    \centering
    \small 
    \setlength{\tabcolsep}{4pt} 
    \renewcommand{\arraystretch}{0.9} 

    \caption{\textbf{{Dataset Information}}} 
    \vspace{0.4cm} 
    \label{tab:dataset_information}
    \begin{tabular}{lccc}
        \hline
        \textbf{Dataset} & \textbf{\#Samples} & \textbf{\#Features} & \textbf{\%Majority} \\
        \hline
        Fertility & 100 & 9 & 88 \\
        Pittsburg-Bridges-T-OR-D & 102 & 7 & 86.3 \\
        Molec-Biol-Promoter & 106 & 57 & 50 \\
        Monks-1 & 124 & 6 & 50 \\
        Echocardiogram & 131 & 10 & 67.2 \\
        Breast-Cancer-Wisc-Prog & 198 & 33 & 76.3 \\
        Parkinsons & 195 & 22 & 75.4 \\
        Statlog-Heart & 270 & 13 & 55.6 \\
        Horse-Colic & 300 & 25 & 63.7 \\
        Musk-1 & 476 & 166 & 56.5 \\
        Ilpd-Indian-Liver & 583 & 9 & 71.4 \\
        Credit-Approval & 690 & 15 & 55.5 \\
        Blood & 748 & 4 & 76.2 \\
        Oocytes-Trisopterus-Nucleus-2f & 912 & 25 & 57.8 \\
        Mammographic & 961 & 5 & 53.7 \\
        Statlog-German-Credit & 1000 & 24 & 70 \\
        Oocytes-Merluccius-Nucleus-4d & 1022 & 41 & 68.7 \\
        titanic & 2201 & 3 & 67.7 \\
        \hline
    \end{tabular}
\end{table}

\subsection{Experimental Setup}

The experiments were conducted on a PC with an Intel Core i5-11320H CPU @ 3.20GHz and 16 GB RAM. The software environment was Google Colab with Python 3.10. We employed \textit{cvx.opt.solvers} to efficiently solve the quadratic programming problem (QQP), ensuring precise optimization for our classification model.
The hyperparameters \(c_1\), \(c_2\), and \(c_u\) were selected from the set \(\{2^{2i} \mid i = -4, \ldots, 4\}\). For simplicity, \(c_1\), \(c_2\), and \(c_u\) were set equal. The model's performance could potentially improve if these parameters were independent, but due to computational limitations, we considered them equal. The values of \(\epsilon\) were considered in the range (0,1) and for simplicity \(\epsilon\) values were selected from set \{0, 0.2, 0.4, 0.6, 0.8, 1\} . 

The values for the minimum number of data points in a granular ball (\textit{num}) and the purity for ball generation (\textit{pur}) were first observed and recorded for each dataset. This preliminary step ensured that favorable ranges for \textit{num} and \textit{pur} were considered when using \(c_1\), \(c_2\), \(c_u\), \textit{num}, and \textit{pur} as hyperparameters. The ball generation program includes a delete ball functionality, which rechecks if the generated balls meet the \textit{num} and \textit{pur} thresholds; otherwise, the balls are deleted. To avoid the cost of hyperparameter tuning in cases where no balls are generated and to select favorable parametric ranges for \textit{num} and \textit{pur} to achieve better accuracy, we observed ball generation beforehand.

Universum data were generated using two methods: (1) initially splitting the data into training, test, and Universum data, then generating Universum data balls from the Universum data, and (2) Splitting the data into only training and test data, generating training data balls, and then generating Universum data balls from the class +1 and class -1 training balls using average centroid and radius parameters. Specifically, we consider an equal number of samples from both class +1 and class -1 (minimum from either class), and then generate Universum data balls using the average centroid from each class +1 and class -1 sample together, along with the corresponding average radii. We conducted both experiments but found the second method to be computationally expensive. Thus, we uniformly followed the first method across all datasets.

To obtain Universum knowledge, we randomly split the data into 50\% training data, 30\% Universum data, and 20\% test data. Universum data granular balls were generated with the same \textit{num} and \textit{pur} thresholds used for training data. We performed 5-fold cross-validation to select the best pair of hyperparameters, and the test data were evaluated using these parameters.

\subsection{Results and Analysis}
Table \ref{tab:accuracy_comparison} presents the accuracy comparison of the models across various datasets.

\begin{table}[h]
    \centering
    \small 
    \setlength{\tabcolsep}{4pt} 
    \renewcommand{\arraystretch}{0.9} 
    \caption{\textbf{Average Accuracy Comparison of Models}}
        \vspace{0.4cm} 

    \label{tab:accuracy_comparison}
    \begin{tabular}{lcccc}
        \hline
        \textbf{Dataset} & \textbf{GBU-TSVM} & \textbf{U-TSVM} & \textbf{TSVM} & \textbf{Pin-GTSVM} \\
        \hline
        Pittsburg-Bridges & \textbf{90.47} & 81.23 & 83.87 & 81.71 \\
        Monks-1 & \textbf{96.87} & 84.38 & 79.27 & 76.31 \\
        Echocardiogram & \textbf{85.18} & 83.54 & 80.33 & 78.66 \\
        Parkinsons & 82.50 & 79.48 & 79.66 & \textbf{83.05} \\
        Statlog-Heart & \textbf{88.83} & 79.06 & 71.26 & 82.71 \\
        Blood & 79.33 & \textbf{80.67} & 64.88 & 76.14 \\
        Ilpd-Indian-Liver & \textbf{75.42} & 72.04 & 66.28 & 60.48 \\
        Credit-Approval & \textbf{85.53} & 81.15 & 85.09 & 81.73 \\
        Mammographic & \textbf{87.56} & 81.69 & 82.01 & 56.74 \\
        Oocytes-Trisopterus & \textbf{83.06} & 77.24 & 79.92 & 73.35 \\
        \hline
    \end{tabular}
\end{table}

It is noteworthy that the GBU-TSVM model consistently outperforms other TSVM variants with increasing dataset sizes. This observation was consistent across various datasets.\\
The accuracy formula used for the experiments is given by:

\small 
\[ \text{Accuracy} = \frac{\text{Number of True Positives + Number of True Negatives}}{\text{Total Number of Predictions}} \times 100\% \]\vspace{0.3cm}
\normalsize 

For example, for datasets with an increasing number of samples (e.g., from 100 to 1000), better results were observed compared to other models.
For the Oocytes-Trisopterus dataset (1022 samples), the average accuracies for the models were as follows: 83.06\% for the GBU-TSVM model, 77.24\% for U-TSVM, 79.92\% for TSVM, and 73.35\% for Pin-GTSVM. Additionally, for datasets with data points in the range of 1000, GBU-TSVM performs 16 times faster than U-TSVM and 58 times faster than TSVM. This can be attributed to the representation of thousands of data points by a few hundred balls or less, leading to reduced computation time. Time for Hyperparameter tuning is excluded.\\

\begin{figure}[h]
  \centering
  \includegraphics[width=0.84\linewidth]{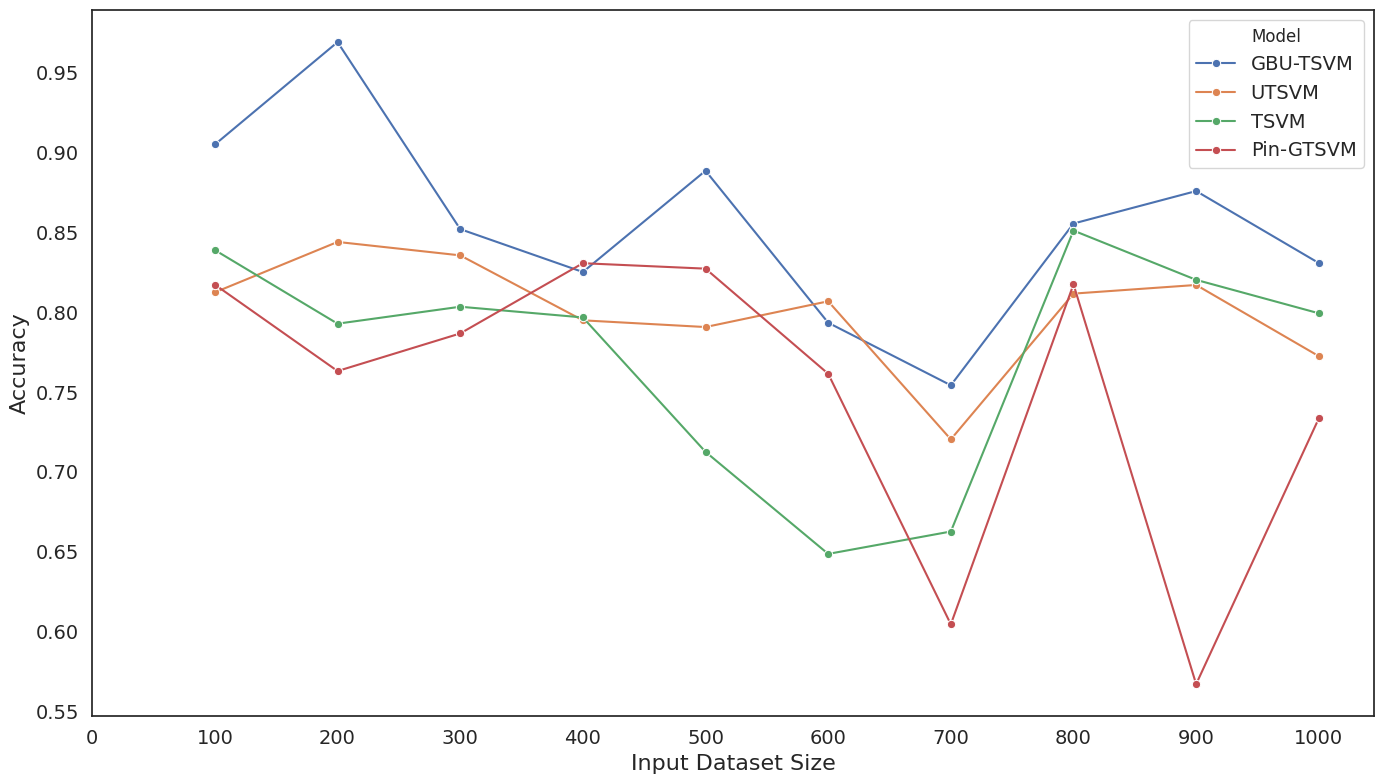}
  \caption{Model Performance Across Various Dataset Sizes}
  \label{fig:performance}
\end{figure}

GBU-TSVM also shows remarkable performance on smaller datasets, especially those with fewer than 300 samples. Please refer to Figure \ref{fig:performance}. The model effectively handles the reduced data size, maintaining high accuracy and computational efficiency. 
\vspace{0.1cm}

All results were obtained through a fair comparison under the same settings. GBU-TSVM is a more generalized version, and U-TSVM, Pin-GTSVM and other TSVM models were derived by making necessary modifications to the GBU-TSVM model to record the accuracies. GBU-TSVM achieves the highest accuracy in 8 out of the 10 datasets, underscoring its consistent superiority across a diverse range of datasets and further validating its robustness. Furthermore, GBU-TSVM not only achieves the highest accuracy in most datasets but also provides a significant accuracy margin over the second-best model. For instance, in the Monks-1 dataset, GBU-TSVM outperforms U-TSVM by a margin of 12.49\%.\\

GBU-TSVM shows strong performance on datasets that are typically challenging, such as the Pittsburg-Bridges and Echocardiogram datasets, where it leads by a notable margin. Moreover, the strong performance of GBU-TSVM on medical datasets like Echocardiogram and Mammographic is noteworthy. Given the importance of accuracy in medical diagnoses, this highlights GBU-TSVM’s potential for critical real-world applications.\\

We have also statistically confirmed that our proposed model, GBY-TSVM, outperforms others. Below are some tests that support this assertion:

\subsubsection{Friedman Test}
\vspace{0.3cm}
The Friedman Test evaluates the statistical significance of differences in accuracy among multiple models. This non-parametric test ranks the models for each dataset and compares the average ranks to determine if there are significant differences.\\
The Friedman Test statistic (\( \chi^2 \)) is calculated as:
\begin{equation}
\chi^2 = \left[ \frac{12}{nk(k+1)} \sum_{i=1}^{k} R_i^2 \right] - 3n(k+1)
\end{equation}
where:
\begin{itemize}
    \item \( n \) is the number of datasets
    \item \( k \) is the number of models
    \item \( R_i \) is the sum of ranks for model \( i \)
\end{itemize}
For our experiments, the Friedman Test yielded the following results:

\begin{table}[h]
    \centering
    \begin{minipage}[t]{0.45\textwidth}
        \centering
        \small
        \setlength{\tabcolsep}{4pt}
        \renewcommand{\arraystretch}{0.9}
        \caption{\textbf{Average Ranks of Models}}
        \vspace{0.4cm}
        \begin{tabular}{@{}ll@{}}
            \toprule
            \textbf{Model} & \textbf{Rank} \\
            \midrule
            GBU-TSVM & 1.2 \\
            U-TSVM & 2.8 \\
            TSVM & 2.8 \\
            Pin-GTSVM & 3.2 \\
            \bottomrule
        \end{tabular}
    \end{minipage}
    \hfill
    \begin{minipage}[t]{0.45\textwidth}
        \centering
        \caption{\textbf{Friedman Test Results}}
        \vspace{0.4cm}
        \begin{tabular}{@{}ll@{}}
            \toprule
            \textbf{$\tilde{\chi}^2$} & 14.16 \\
            \textbf{P-value} & 0.0027 \\
            \bottomrule
        \end{tabular}
    \end{minipage}
\end{table}
The p-value of 0.0027 suggests a statistically significant difference in model performance. With this p-value being less than the chosen significance level (typically 0.05), the null hypothesis is rejected, indicating significant differences in accuracy among the models.

\subsubsection{Wilcoxon Signed-Rank Test}
\vspace{0.30cm}

The Wilcoxon Signed-Rank Test was performed to compare the performance of GBU-TSVM against each of the other models pairwise. This non-parametric test is used to compare two related samples to assess whether their population mean ranks differ.\\

The Wilcoxon Signed-Rank Test statistic is calculated as follows:
\begin{equation}
    W = \sum_{i=1}^{n} \text{sign}(d_i) \cdot R_i    
\end{equation}
where:
\begin{itemize}
    \item \( d_i \) is the difference between each pair of observations.
    \item \( \text{sign}(d_i) \) is the sign of the difference (\(+1\) if positive, \(-1\) if negative).
    \item \( R_i \) is the rank of the absolute value of \( d_i \).
    \item \( n \) is the number of pairs.
\end{itemize}
The null hypothesis (\( H_0 \)) assumes that there is no difference between the paired observations, while the alternative hypothesis (\( H_1 \)) suggests that there is a significant difference.\\
For our experiments, the results are presented in the table below:

\begin{table}[h]
    \centering
    \setlength{\tabcolsep}{4pt}
    \renewcommand{\arraystretch}{0.9}
    \caption{\textbf{Wilcoxon Signed-Rank Test Results}}
    \label{tab:wilcoxon_test}
    \vspace{0.4cm}
    \begin{tabular}{lcc}
        \toprule
        \textbf{Comparison} & \textbf{p-value} & \textbf{Conclusion} \\
        \midrule
        GBU-TSVM vs. U-TSVM & 0.0039 & Significant \\
        GBU-TSVM vs. TSVM & 0.0019 & Significant \\
        GBU-TSVM vs. Pin-GTSVM & 0.0039 & Significant \\
        \bottomrule
    \end{tabular}
    \\
\end{table}
The p-values indicate that there are significant differences in performance between GBU-TSVM and each of the other models, using a significance threshold of \( \alpha = 0.05 \).

\subsubsection{Kruskal-Wallis Test}
\vspace{0.3cm}
The Kruskal-Wallis Test is performed to compare the performance of multiple TSVM variants across multiple datasets. This non-parametric test assesses whether there are significant differences in performance among the TSVM variants.\\
The Kruskal-Wallis test statistic is calculated as follows:
\begin{equation}
    H = \frac{12}{N(N+1)} \left[ \sum_{j=1}^{k} \frac{R_j^2}{n_j} - 3(N+1) \right]
\end{equation}
where:
\begin{itemize}
    \item \( H \) is the Kruskal-Wallis test statistic.
    \item \( N \) is the total number of observations across all groups.
    \item \( k \) is the number of groups (TSVM variants).
    \item \( R_j \) is the sum of ranks for the \( j \)-th group.
    \item \( n_j \) is the number of observations in the \( j \)-th group.
\end{itemize}


\begin{table}[h]
    \centering
    \setlength{\tabcolsep}{4pt}
    \renewcommand{\arraystretch}{1}
    \caption{\textbf{Results of Kruskal-Wallis Test}}
    \vspace{0.4cm}
    \label{tab:kruskal_wallis_results}
    \begin{tabular}{lc}
        \hline
        \textbf{Statistic} & \textbf{Value} \\
        \hline
        test statistic & 10.63 \\
        p-value & 0.0139 \\
        \hline
    \end{tabular}
\end{table}

With a significance level of $\alpha = 0.05$, since the p-value (0.0139) is less than $\alpha$, we reject the null hypothesis. Therefore, we conclude that there are significant differences in performance among the TSVM variants.\\
This implies that at least one of the TSVM variants performs significantly differently compared to the others across the datasets considered.

\subsubsection{Win-Tie-Loss Analysis}
\vspace{0.3cm}
To further compare the performance of the models, we conducted a Win-Tie-Loss analysis. It involves counting the number of times GBU-TSVM outperforms (wins), ties with, or is outperformed by (losses) each of the other models across multiple datasets.

\begin{table}[h]
    \centering
    \setlength{\tabcolsep}{4pt}
    \renewcommand{\arraystretch}{1}
    \caption{\textbf{Win-Tie-Loss Analysis}}
    \vspace{0.4cm}
    \label{tab:win_tie_loss}
    \begin{tabular}{lccc}
        \hline
        \textbf{Comparison} & \textbf{Wins} & \textbf{Ties} & \textbf{Losses} \\
        \hline
        GBU-TSVM vs. U-TSVM & 9 & 0 & 1 \\
        GBU-TSVM vs. TSVM & 9 & 1 & 0 \\
        GBU-TSVM vs. Pin-GTSVM & 9 & 0 & 1 \\
        \hline
    \end{tabular}
\end{table}
These results demonstrate that GBU-TSVM consistently outperforms U-TSVM, TSVM, and Pin-GTSVM in terms of accuracy across the evaluated datasets, with no ties and only a few losses in each comparison.

\subsubsection{Parameter Sensitivity Analysis}
\vspace{0.3cm}
We conducted a detailed parameter sensitivity analysis to understand the impact of different hyperparameter settings on the performance of GBU-TSVM. Specifically, we varied the parameters \(c_1\), \(c_2\), \(c_u\), and \(\epsilon\), and recorded their effects on accuracy. The analysis was carried out on the Statlog-Heart dataset, which consists of 270 samples and 13 features, with a majority class proportion of 55.6\%.

For our experiments, we first observed the granular ball formation and determined that a minimum number of data points in a granular ball (\(\text{num}\)) of 20 and a purity of a granular ball (\(\text{pur}\)) of 0.88 were suitable settings. Therefore, we fixed \(\text{num}\) at 20 and \(\text{pur}\) at 0.88. The hyperparameters \(c_1\), \(c_2\), \(c_u\), and \(\epsilon\) were varied across a range defined by powers of 2 from \(2^{-8}\) to \(2^{8}\):
\[
\{2^k \mid k \in \{-8, -7, \ldots, 7, 8\}\}
\]
This range allowed us to comprehensively explore the parameter space and identify the most effective settings.


The tables below summarize the top 1\% accuracy results for each parameter, highlighting the top value for \(c_1\), \(c_2\), \(c_u\), and \(\epsilon\).

\begin{table}[h]
    \centering
    \begin{tabular}{|c|c|}
        \hline
        \textbf{Parameter} & \textbf{Value} \\
        \hline
        \(c_1\) & 8.0 \\
        \(c_2\) & 4.0 \\
        \(c_u\) & 0.25 \\
        \(\epsilon\) & 32.0 \\
        \hline
    \end{tabular}
    \caption{Top Values for Each Parameter}
    \label{tab:top_values}
\end{table}

The hyperparameters \(c_1 = 8.0\), \(c_2 = 4.0\), \(c_u = 0.25\), and \(\epsilon = 32.0\) were identified as the most effective, emphasizing the critical role of optimal hyperparameter selection in achieving peak classification accuracy. These findings underscore the necessity of finely tuning hyperparameters to optimize GBU-TSVM's performance. While the model exhibits robustness across a range of values, pinpointing and configuring the most effective parameters are crucial steps toward maximizing classification accuracy.

\section{Conclusion} \label{sec:conclusion}
In this paper, we proposed the Granular Ball-based Universum Twin Support Vector Machine (GBU-TSVM), leveraging granular-ball computing for enhanced classification. Our model integrates traditional TSVM methodologies with granular-ball learning to effectively utilize Universum data. This approach allows for the incorporation of unlabeled samples, enriching the classifier's performance by embedding prior domain knowledge.
The experimental results on various UCI benchmark datasets demonstrate that GBU-TSVM achieves superior performance compared to traditional TSVMs and it's variants, particularly in handling imbalanced and complex datasets. The incorporation of granular-ball structures significantly enhances the classifier's robustness and accuracy. In conclusion, the proposed GBU-TSVM model provides a powerful framework for classification tasks, offering improved generalization capabilities and resilience to data variations. Future work could explore the optimization of hyperparameters independently and extend the application of GBU-TSVM to multi-class classification problems.

\phantomsection
\label{sec:References}

\bibliographystyle{unsrt}
\bibliography{references}

\end{document}